\documentclass{article}

\PassOptionsToPackage{numbers, compress}{natbib}

    \usepackage[preprint]{neurips_2024}

\usepackage[utf8]{inputenc} 
\usepackage[T1]{fontenc}    
\usepackage{hyperref}       
\usepackage{url}            
\usepackage{booktabs}       
\usepackage{amsfonts}       
\usepackage{nicefrac}       
\usepackage{microtype}      
\usepackage{xcolor}         

\usepackage{kotex}
\usepackage{graphicx}
\usepackage{caption}
\usepackage{subcaption}
\usepackage{multirow}
\usepackage{amsmath}
\usepackage{algorithm}
\usepackage{algpseudocode}

\usepackage{wrapfig}
\title{Realizing Unaligned Block-wise Pruning for DNN Acceleration on Mobile Devices}

\author{%
  Hayun Lee \\
  Sungkyunkwan University\\
  \texttt{lhy920806@skku.edu} \\
  \And
  Dongkun Shin\thanks{Corresponding author.} \\
  Sungkyunkwan University\\
  \texttt{dongkun@skku.edu} \\
}

\begin{document}

\maketitle

\begin{abstract}
With the recent proliferation of on-device AI, there is an increasing need to run computationally intensive DNNs directly on mobile devices. However, the limited computing and memory resources of these devices necessitate effective pruning techniques. Block-wise pruning is promising due to its low accuracy drop tradeoff for speedup gains, but it requires block positions to be aligned with block size, hindering optimal position selection to minimize model accuracy drop. Unaligned block pruning (UBP) addresses this by allowing blocks to be selected at arbitrary positions, yet its practical use is limited by a time-consuming optimal block selection algorithm and lack of efficient inference kernels. In this paper, we propose a pseudo-optimal yet fast block selection algorithm called Block Expansion and Division (BED), which can be integrated into an iterative model training process. Additionally, we introduce an efficient inference kernel implementation for mobile devices, enabling a UBP-based model to achieve similar latency to a DNN model compressed by aligned block pruning. We demonstrate the superiority of our techniques on a real mobile phone with MobileNet and ResNet models.
\end{abstract}

\section{Introduction}

Deep neural networks (DNNs) have emerged as powerful machine learning techniques, demonstrating exceptional performance across various applications. Mobile devices particularly benefit from DNN applications due to their ability to process user-generated content such as photos, videos, voice recordings, and data from various sensors. Modern mobile devices increasingly incorporate AI functionalities like scene detection and image noise reduction~\cite{ignatov2021fast, ignatov2021fast2, ignatov2022learned}.

However, DNN models often require substantial computational resources and memory, posing significant challenges for mobile devices with limited capabilities. Although cloud computing presents a potential solution, it raises concerns regarding privacy, network latency, and connectivity dependence. Consequently, there is a growing interest in deploying DNNs directly on mobile devices, known as on-device AI. To support this trend, manufacturers are developing AI accelerators like GPUs or NPUs and optimizing DNN technologies. Despite these advancements, scenarios still exist where only CPU acceleration is available on mobile devices.

Extensive research focuses on optimizing DNN models for mobile deployment, leading to the development of mobile-friendly, compact models such as MobileNet~\cite{mobilenet, mobilenetv2, mobilenetv3}, which deliver respectable performance with reduced computational demand and memory usage. Model compression techniques such as network pruning~\cite{network_slimming, thinet, amc, han2015learning, dai2019nest, fast, ovw, bcbp, patdnn, nm_sparsity} and quantization~\cite{pact, qil, lsq} have been explored to further decrease these demands while maintaining accuracy.

\begin{figure}
  \centering
  \includegraphics[width=0.8\linewidth]{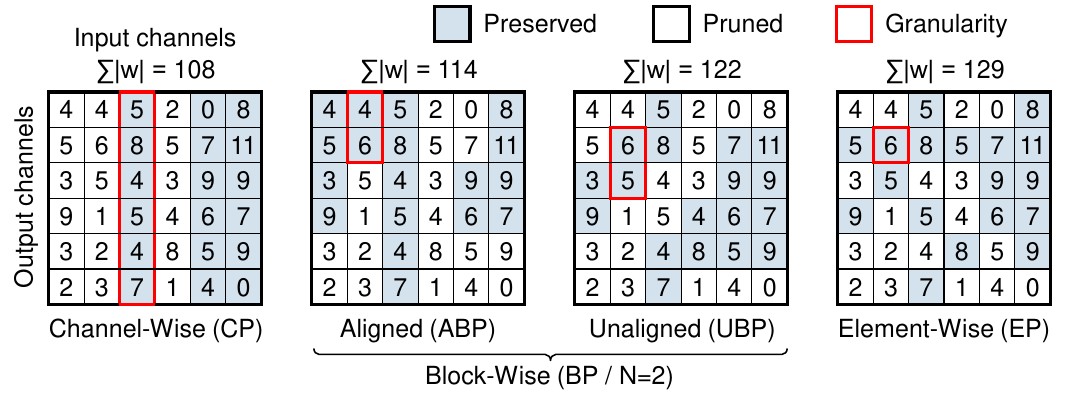}
  \vspace{-5pt}
  \caption{Various pruning patterns.}
  \label{fig:sparsity_pattern}
  \vspace{-20pt}
\end{figure}

Given the inherent redundancy in neural network models, pruning plays a crucial role by eliminating non-essential weight parameters, thus transforming dense networks into sparse ones. This transformation allows networks to focus only on significant weights, enhancing performance efficiency. Pruning can be categorized into various sparse patterns based on granularity: channel~\cite{network_slimming, thinet, amc}, block~\cite{fast, ovw, bcbp, nm_sparsity}, and element~\cite{han2015learning, dai2019nest} levels. Figure~\ref{fig:sparsity_pattern} illustrates the sum of importance scores of preserved weights at a target sparsity of 50\% using the $\ell_1$ norm as the pruning criterion. Channel-wise pruning (CP) creates structured sparse patterns that enhance hardware efficiency but may compromise accuracy. Conversely, element-wise pruning (EP) maintains accuracy by selecting important weights with fine granularity, which, however, might reduce hardware efficiency due to irregular data access patterns.

Block-wise Pruning (BP), which minimizes accuracy impact while being hardware-friendly, has recently gained significant attention. BP involves grouping consecutive kernels or elements for pruning. Compared to CP, BP minimizes accuracy loss, and compared to EP, it provides higher hardware efficiency, presenting an advantageous trade-off among various sparsity patterns. BP can be further categorized into aligned block-wise pruning (ABP)~\cite{1xn, ovw, shfl-bw, bcbp, fast, SUBP} and unaligned block-wise pruning (UBP)~\cite{unaligned} based on constraints on block positions. ABP requires blocks to be aligned according to block size, whereas UBP permits block selection at arbitrary positions, potentially achieving higher importance scores. 

However, the practical implementation of UBP faces challenges, notably the complexity of block selection and the lack of efficient inference kernels. This paper provides a detailed discussion of these challenges and proposes methods to overcome them. By doing so, we aim to make UBP a viable technology for use on mobile devices.

In this work, we make the following contributions:
\begin{itemize}
    \item We introduce the Block Expansion and Division (BED) algorithm for unaligned block-wise pruning, which offers a fast and effective block selection mechanism. BED achieves pseudo-optimal performance with a search speed comparable to greedy approaches, enhancing the efficiency and scalability of pruning processes.
    \item We propose a Weight Rotating and Output Stationary (WROS) dataflow that effectively eliminates the additional overhead associated with processing unaligned blocks. This dataflow enables UBP kernels to achieve performance nearly identical to that of ABP kernels.
    \item We empirically validate the superiority of our methods through experiments on real mobile devices using MobileNet and ResNet architectures. Our results demonstrate that UBP implemented with our method achieves higher accuracy than ABP, while maintaining comparable latency levels.
\end{itemize}

\section{Related Work}

\paragraph{Pruning.}
Pruning techniques in neural networks are essential for reducing model size and computational complexity, which is crucial for deployment on resource-constrained devices. These techniques are typically categorized into channel-wise, element-wise, and block-wise pruning, each offering distinct trade-offs between efficiency and accuracy.
\textbf{Channel-wise pruning}~\cite{network_slimming, thinet, amc} removes entire channels from the network. This method maintains the convolutional structure, facilitating the continued use of efficient inference libraries. However, channel-wise pruning may significantly impact accuracy due to the coarse granularity of the reductions.
\textbf{Element-wise pruning}~\cite{han2015learning, dai2019nest} targets individual weights to create fine-grained sparsity. This approach preserves accuracy even at high sparsity levels, but the resulting irregular memory access patterns often lead to only modest latency improvements.
\textbf{Block-wise pruning}~\cite{1xn, SUBP, unaligned} offers a compromise, pruning structured blocks of weights that are smaller than channels but larger than individual elements. This method effectively balances minimizing accuracy loss with enhanced latency improvements through better data reuse and vector instruction utilization. For example, the 1$\times$N~\cite{1xn} pruning pattern specifies a block using consecutive output kernels within the same input channel index. Block-wise pruning can be subdivided into aligned block-wise pruning (ABP)~\cite{1xn, SUBP} and unaligned block-wise pruning (UBP)~\cite{unaligned}. Although UBP can potentially achieve higher accuracy, it has been historically challenging to implement due to a time-consuming block selection process. We propose a novel block selection method that is both time-efficient and near-optimal, thereby making the training of UBP more practical.

\paragraph{Sparse DNN Inference Engines for Mobile CPUs.}
Efficient inference of sparse networks on mobile CPUs requires specialized software frameworks. XNNPACK~\cite{xnnpack} provides highly optimized implementations for neural network operations, including SpMM kernels that support sparse inference for 1x1 convolutions. TVM~\cite{tvm} offers a comprehensive optimization stack that enables effective code generation for 1$\times$N pruning patterns. Currently, most sparse DNN inference engines focus on ABP, with no kernels specifically designed for UBP. This paper addresses the challenges associated with developing UBP kernels and proposes a novel dataflow that achieves performance levels comparable to those of existing ABP kernels.

\section{Unaligned Block-wise Pruning and Challenges}

In this section, we establish the 1$\times$N pruning pattern as the baseline for our study on Aligned Block-wise Pruning (ABP). This pattern takes full advantage of the NEON SIMD instruction set provided by the ARM architecture, facilitating efficient computational operations. By maintaining contiguous non-zero weights, the 1$\times$N pattern aligns with the parallel processing capabilities of SIMD, enabling efficient vector operations that are crucial for mobile computing.

Building on this, we define Unaligned Block-wise Pruning (UBP), an extension of the 1$\times$N pattern. UBP aims to harness the computational efficiency of SIMD while potentially improving model accuracy by allowing more flexibility in the placement of non-zero weights across the network.

Following this, we will introduce the challenges associated with UBP. These include the complexity of the block selection algorithm, which can significantly prolong the training process, and the practical difficulties in deploying efficient inference kernels, which can limit UBP's applicability in real-world scenarios.

\subsection{Unaligned 1$\times$N Block-wise Pruning}

Unaligned Block-wise Pruning (UBP) adapts the 1$\times$N pruning pattern, where one block in the convolution layer weights is defined as a sequence of $N$ consecutive output kernels sharing the same input channel index. For each convolution layer $l$, with weights $\mathbf{W}^l \in \mathbb{R}^{c_{l+1} \times c_l \times h_l \times w_l}$, blocks of size $(N, 1, h_l, w_l)$ are selected based on the pruning rate $p_l$, and weights outside these blocks are pruned. Here, $c_l$, $c_{l+1}$, $h_l$, and $w_l$ denote the input channels, output channels, kernel height, and kernel width of layer $l$, respectively.

The starting position of a block can be any output channel and input channel, allowing for a theoretical number of blocks per layer $b_l = c_{l+1} \times c_l$. The number of selected blocks is calculated as $m_l = \left\lfloor{b_l \cdot (1 - p_l) / N}\right\rfloor$. Each block starting at the $i$-th output channel and the $j$-th input channel is indexed as $k$ (where $k = i + c_{l+1} \cdot j$). The importance score $\mathbf{S}^l_k$ of the $k$-th block is computed as follows:
\vspace{-5pt}
\begin{equation}
    \mathbf{S}^l_{k} = \begin{cases}
        \mathcal{F}\left( \mathbf{W}^l_{i:i+N, j, :, :} \right) & \text{if } i + N - 1 < c_{l+1} \\
        -\infty & \text{otherwise}
    \end{cases}
\end{equation}
where $\mathcal{F}(\cdot)$ is the function used to calculate the importance score, typically the $\ell_1$ norm in this study. If blocks that exceed the boundaries of the weights are not eligible for selection (e.g., starting output channel index $i$ exceeds $c_{l+1}-N$), the score is set to $-\infty$ to exclude these blocks in subsequent calculations.

The mask $\mathbf{M}^l_k \in \{0, 1\}$ for the $k$-th block is set to 1 when the block is selected and 0 otherwise. This mask ensures that only up to one out of every consecutive $N$ blocks is selected, preventing overlap between chosen blocks and meeting the pruning rate. Thus, the objective function for UBP is defined as:
\vspace{-5pt}
\begin{equation}\label{eq:bw-objective}
    \operatorname*{arg\,max}_{\mathbf{M}^l}{\sum_{l=1}^{L}{\sum_{j}{\mathbf{S}^{l}_{k} \cdot \mathbf{M}^{l}_{k}}}}, \quad \text{s.t.} \quad \sum_k{\mathbf{M}^l_k} = m_l, \quad \sum_{k=n}^{n+N-1}{M^l_k} \in \{0, 1\}, \quad \forall 0 \leq n < b_l - N
\end{equation}
In this study, for the purposes of experimentation, all values of $p_l$ are set uniformly across layers. Notably, if during score calculation the starting output channel index of a block is not a multiple of $N$, setting the score to $-\infty$ effectively converts the method to Aligned Block-wise Pruning (ABP).

\subsection{Challenge 1: Time-consuming Optimal Block Selection Algorithm}

\paragraph{Greedy block selection.}
The simplest method for block selection in block-wise pruning is the greedy approach, where blocks are sorted by their importance scores $\mathbf{S}^l$ and the highest scoring blocks are selected first. The time complexity of this method is primarily determined by the sorting process, which is $O(b_l \log b_l)$. In UBP, when a $k$-th block is selected, the scores of any overlapping blocks are set to $-\infty$ to prevent their selection in subsequent steps:
\begin{equation}
    \mathbf{S}^l_{k+n} = -\infty, \quad \forall -N < n < N
\end{equation}
Although slower than ABP's block selection, where selecting one block does not affect the selection of others, this process can still be executed relatively quickly. However, it inherently limits the ability to find the optimal solution because selected blocks generate non-selectable regions, as defined in Eq.~\ref{eq:bw-objective}.

\begin{wrapfigure}{r}{0.33\textwidth}
    \centering
    \begin{subfigure}[t]{0.2\textwidth}
        \centering
        \includegraphics[height=3cm]{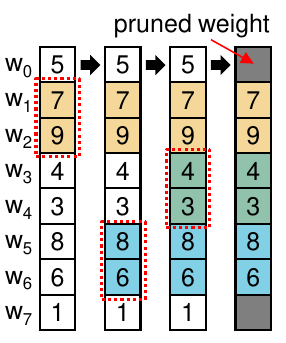}
        \vspace{-5pt}
        \caption{Greedy.}
        \label{fig:greedy-vs}
    \end{subfigure}
    \hfill
    \begin{subfigure}[t]{0.12\textwidth}
        \centering
        \includegraphics[height=3cm]{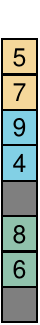}
        \vspace{-5pt}
        \caption{Optimal solution.}
        \label{fig:optimal-vs}
    \end{subfigure}
    \caption{Greedy block selection examples in UBP.}
    \label{fig:vs}
\end{wrapfigure}

Fig.~\ref{fig:greedy-vs} illustrates an example of greedy block selection in UBP where the block size is 2 and the target sparsity is 25\%. The numbers in each element represent their importance scores. The process starts with a dense model, and the most important weight block is sequentially chosen until the target sparsity is achieved in the order of $(w_1, w_2)$, $(w_5, w_6)$, and $(w_3, w_4)$. In UBP, the starting location of a selected block can be any position, and only contiguous blocks not overlapping with previously selected blocks are considered. Once the target number of blocks is reached, the remaining weights ($w_0$ and $w_7$) are discarded. However, this greedy approach has limitations in finding the optimal solution (see Fig.~\ref{fig:optimal-vs}) due to the potential overlap of candidate blocks. For example, once the block $(w_1, w_2)$ is selected, the block $(w_0, w_1)$ becomes ineligible for selection. Consequently, weight $w_0$ is pruned, although weight $w_4$, which has a lower importance score, is selected.

\paragraph{Optimal block selection.}
A recently proposed dynamic programming-based optimal block selection algorithm~\cite{unaligned} considers all candidate blocks to find the optimal solution for UBP. When optimally selecting $n$ blocks out of the first $k$, the importance score and block index set are denoted as $\mathbf{T}^l_{k, n}$ and $\mathbf{I}^l_{k, n}$, respectively. To determine $\mathbf{T}^l_{k, n}$, one must decide whether selecting or not selecting the $k$-th block is beneficial. If the $k$-th block is selected, the optimal score is obtained by adding the score of the $k$-th block $\mathbf{S}^l_k$ to the optimal score of selecting $n-1$ blocks out of the $k-N$ non-overlapping blocks, $\mathbf{T}^l_{k-N, n-1}$. Conversely, if the $k$-th block is not selected, the optimal score is the same as the optimal score of selecting $n$ blocks from the first $k-1$ blocks, $\mathbf{T}^l_{k-1, n}$. The index set $\mathbf{I}^l_{k, n}$ that satisfies $\mathbf{S}^l_k$ is updated accordingly:
\vspace{-5pt}
\begin{equation}
    \mathbf{T}^l_{k, n} = \max \left( \mathbf{T}^l_{k-N, n-1} + \mathbf{S}^l_k, \mathbf{T}^l_{k-1, n} \right), \quad
    \mathbf{I}^l_{k, n} = \begin{cases} 
        \mathbf{I}^l_{k-N, n-1} \cup \{k\} & \text{if } \mathbf{T}^l_{k, n} > \mathbf{T}^l_{k-1, n} \\ 
        \mathbf{I}^l_{k-1, n} & \text{otherwise}
    \end{cases}
\end{equation}
\vspace{-5pt}

The ultimate goal is to determine the optimal set of block indices $\mathbf{I}^l_{b_l, m_l}$ for selecting $m_l$ blocks out of $b_l$. Due to the iterative nature of this process, the time complexity is $O(b_l \cdot m_l^2)$, and as the number of blocks increases, the computation time grows exponentially, making it impractical for training. This complexity contrasts with gradual pruning~\cite{zhu2017prune}, ADMM~\cite{admmnn, patdnn}, and Grow-and-Prune~\cite{grow-and-prune} techniques, which require continuous block selection unlike one-shot pruning. For instance, pruning a MobileNetV1 model with $N=4$ and a target sparsity of 90\% on the ImageNet dataset takes about 20 minutes per epoch on a single RTX 3090 GPU, while optimal block selection can take several hours. This lengthy process leads to inefficient training as the GPU remains largely idle. Thus, for practical UBP training, a block selection method that balances speed and near-optimal performance is essential.

\subsection{Challenge 2: Need for Efficient Inference Kernels}

\begin{figure}
  \centering
  \includegraphics[width=\textwidth]{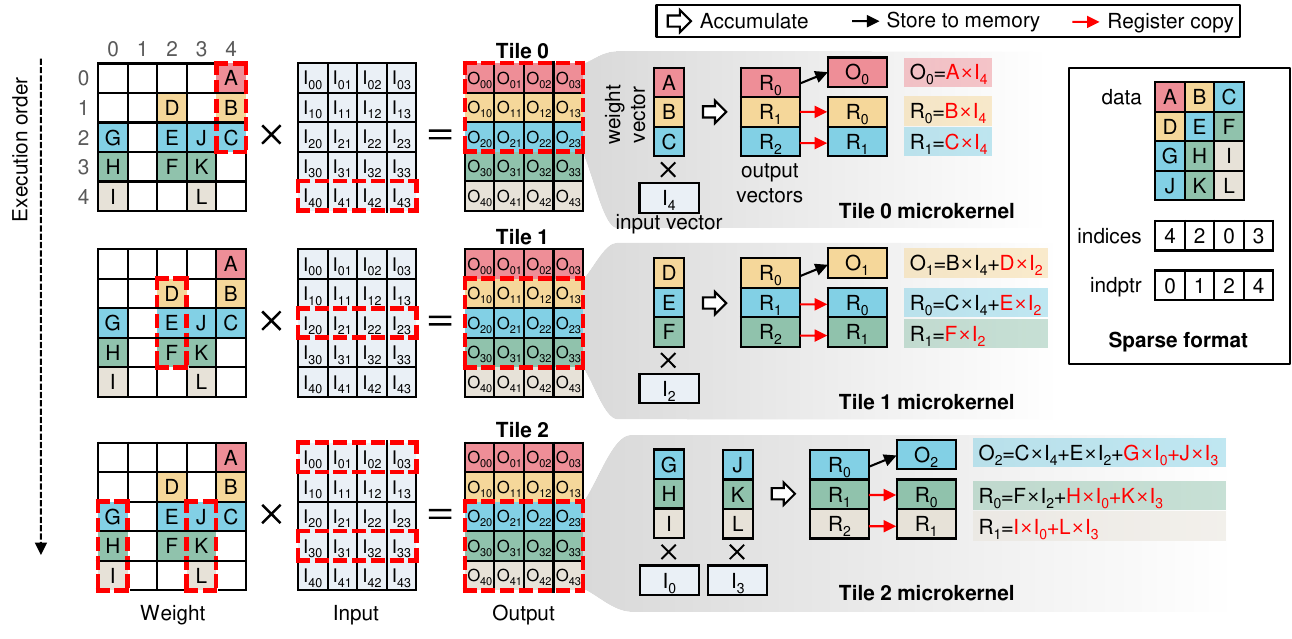}
  \vspace{-10pt}
  \caption{Overlapped output tiles problem in UBP kernel and na\"ive implementation.}
  \label{fig:motive-kernel}
  \vspace{-15pt}
\end{figure}

\paragraph{Overlapped Output Tiles Problem.}
In computational kernels, tiling is commonly used to efficiently leverage data locality, where increasing the size of tiles can improve performance due to enhanced data reuse. In block-wise pruning, aligning the tile size with the block size can further boost kernel performance. However, Unaligned Block Pruning (UBP) introduces the challenge of overlapping output tiles, an issue that does not occur with Aligned Block Pruning (ABP).

Fig.~\ref{fig:motive-kernel} illustrates the overlapping output tiles that occur during the execution of a UBP kernel in a 1x1 convolution layer with a kernel size of 1 and $N=3$. In this example, tiling is conducted using a 3$\times$4 grid, and the red dotted boxes highlight the data accessed during each tile's computation. In conventional ABP kernels, the row index of output tiles is incremented by $N$, avoiding overlaps. In contrast, UBP kernels increment the row index by 1, leading to overlapping outputs between tiles. For instance, to compute the result for $O_{23}$, accumulations are required from $C \times I_{43}$ in Tile 0, $E \times I_{23}$ in Tile 1, and both $G \times I_{03} + J \times I_{33}$ in Tile 2. Although UBP can achieve computational efficiency comparable to ABP with identical output tile sizes, differences in how outputs are stored in memory create distinct challenges.

\paragraph{Na\"ive Implementation.}
In the basic implementation of Unaligned Block-wise Pruning (UBP), only the first row's output vector is saved in memory after each tile, with subsequent vectors shifted among registers. For example, post Tile 0 processing, only $R_0$ is stored at memory location $O_0$, while results in $R_1$ and $R_2$ are moved to $R_0$ and $R_1$. This shifting process, repeated for each tile, leads to increased overhead from frequent register transfers, particularly as block size and target sparsity grow. This results in performance inefficiencies compared to ABP kernels. Therefore, optimizing UBP requires developing an inference kernel that retains ABP's tile size but minimizes these overheads.

\section{Methodology for Realizing Unaligned Block-wise Pruning}

\subsection{BED: Block Expansion and Division for Training}

\begin{figure}
    \centering
    \begin{minipage}[t]{0.54\textwidth}
        \centering
        \begin{algorithm}[H]
        \small
        \caption{Block expansion algorithm}
        \label{alg:expansion}
        \begin{algorithmic}[1]
            \Procedure{BlockExpansion}{$\mathbf{S}^l, m_l$}
                \State $b_l \gets |\mathbf{S}^l|$
                \State $\mathbf{R}^l \gets \{k \mid 0 \leq k < b_l\}$
                \State $\mathbf{\bar{I}}^l \gets \emptyset$
                \For{$i \gets 1 \text{ to } m_l$}
                    \State $k \gets \operatorname*{arg\,max}_{j}{\mathbf{S}^l_j}$
                    \For{$n \gets 1 \text{ to } N-1$}
                        \State $\mathbf{S}^l_{k-n} \gets \mathbf{S}^l_{k-n} + \mathbf{S}^l_{k-n+N} - \mathbf{S}^l_{k}$
                    \EndFor
                    \State $\mathbf{S}^l \gets \mathbf{S}^l \setminus \{\mathbf{S}^l_{k}, \mathbf{S}^l_{k+1}, \ldots, \mathbf{S}^l_{k+N-1}\}$
                    \State $\mathbf{\bar{I}}^l = \mathbf{\bar{I}}^I \cup \mathbf{R}^l_k$
                    \State $\mathbf{R}^l = \mathbf{R}^l \setminus \{\mathbf{R}^l_{k}, \mathbf{R}^l_{k+1}, \ldots, \mathbf{R}^l_{k+N-1}\}$
                \EndFor
                \State \Return $\mathbf{\bar{I}}^l$
            \EndProcedure
        \end{algorithmic}
        \end{algorithm}
    \end{minipage}
    \hfill
    \begin{minipage}[t]{0.45\textwidth}
        \centering
        \begin{algorithm}[H]
        \small
        \caption{Block division algorithm}
        \label{alg:division}
        \begin{algorithmic}[1]
            \Procedure{BlockDivision}{$\mathbf{\bar{I}}^l, N$}
                \State $\mathbf{\bar{I}}^l \gets \text{sort}(\mathbf{\bar{I}}^l)$
                \State $\mathbf{I}^l \gets \emptyset$
                \State $next\_idx \gets 0$
                \ForAll{$k \in \mathbf{\bar{I}}^l$}
                    \If{$next\_idx > k$}
                        \State $\mathbf{I}^l \gets \mathbf{I}^l \cup \{next\_idx\}$
                        \State $next\_idx \gets next\_idx + N$
                    \Else
                        \State $\mathbf{I}^l \gets \mathbf{I}^l \cup \{k\}$
                        \State $next\_idx \gets k + N$
                    \EndIf
                \EndFor
                \State \Return $\mathbf{I}^l$
            \EndProcedure
        \end{algorithmic}
        \end{algorithm}
    \end{minipage}
    \vspace{-10pt}
\end{figure}

To overcome the limitations of greedy block selection, we propose the Block Expansion and Division (BED) method. Our BED method allows for the expansion of a selected block by including adjacent elements, ensuring that the total number of elements added equals the target block size. Consequently, the size of the expanded block is always a multiple of the target block size, facilitating its subdivision into blocks of the original size during the final reorganization phase. This expansion process effectively rescues blocks that would otherwise be excluded due to their overlap with a previously selected block.

Algorithm~\ref{alg:expansion} details the expansion process. This involves expanding blocks based on importance scores and generating an index set $\mathbf{\bar{I}}^l$ from the selected indices. While this method resembles the greedy block selection algorithm, it differs in its handling of the scores of adjacent blocks. Unlike the greedy approach, which sets the scores of all overlapping blocks to $-\infty$ to prevent their selection, BED recalculates the block score assuming no weight kernels belong to the $k$-th block (line 8) and excludes the scores of blocks starting from those weight kernels from the score set (line 10). The index of the $k$-th block is then added to the index set $\mathbf{\bar{I}}^l$. As elements are continuously removed from the score set $\mathbf{S}^l$, an index set $\mathbf{R}$ is maintained for the remaining blocks. The algorithm terminates after selecting $m_l$ blocks, returning the final index set $\mathbf{\bar{I}}^l$.

\begin{wrapfigure}{r}{0.2\textwidth}
    \centering
    \includegraphics[height=3cm]{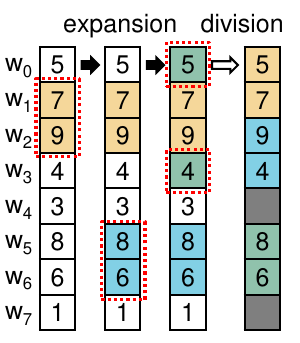}
    \vspace{-5pt}
    \caption{Example of block expansion and division.}
    \label{fig:ved-example}
\end{wrapfigure}

The indices in $\mathbf{\bar{I}}$ are not final and require reconfiguration through the division process, as described in Algorithm~\ref{alg:division}. This algorithm starts by sorting the indices in $\mathbf{\bar{I}}$. Then, if the next expected block index $next\_idx$ is greater than the index $k$ from $\mathbf{\bar{I}}^l$, it indicates an expanded block, and $next\_idx$ is added to the final index set $\mathbf{I}^l$ and increased by $N$ (lines 6-8). If not, it signals the appearance of a new block, and the index $k$ for this new block is added to $\mathbf{I}^l$, setting the next expected block index $next\_idx$ to $k+N$. The process continues until all indices are processed, resulting in the final index set $\mathbf{I}^l$.

An example is illustrated in Fig.~\ref{fig:ved-example}, where weights $w_0$ and $w_3$ are merged into the selected block $(w_1, w_2)$ because their combined importance scores exceed that of the block $(w_3, w_4)$. Following the selection, the combined block $(w_0, w_1, w_2, w_3)$ is divided into $(w_0, w_1)$ and $(w_2, w_3)$.
The BED algorithm iterates $m_l$ times to identify the block with the highest importance score among all $b_l$ blocks, resulting in a time complexity of $O(b_l \cdot m_l)$. While slower than greedy block selection, it is considerably faster than optimal block selection, making it a practical and efficient approach.

\subsection{WROS: Weight Rotating and Output Stationary Dataflow for Inference}

\begin{figure}
    \centering
    \begin{subfigure}{0.33\textwidth}
        \includegraphics[width=\textwidth]{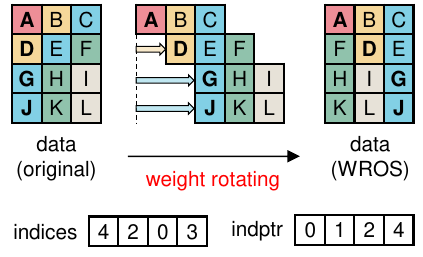}
        \vspace{-10pt}
        \caption{Sparse weight format for WROS.}
        \label{fig:wros-weight}
    \end{subfigure}
    \hfill
    \begin{subfigure}{0.65\textwidth}
        \includegraphics[width=\textwidth]{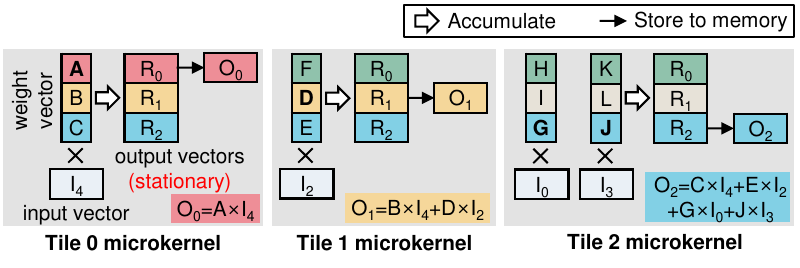}
        \vspace{-10pt}
        \caption{Microkernel execution example in WROS dataflow.}
        \label{fig:wros-kernel}
    \end{subfigure}

    \caption{Sparse format and microkernel execution for weight rotating and output stationary dataflow.}
    \label{fig:wros}
    \vspace{-10pt}
\end{figure}

To address the overlapped output tile problem in UBP kernels, we propose the Weight Rotating and Output Stationary (WROS) dataflow. Unlike the naive approach depicted in Fig.\ref{fig:motive-kernel}, where output register file values are rotated at the end of each microkernel, the WROS dataflow rotates the elements of the weight block in the opposite direction. This rotation maintains the stationary state of output register file values during computation. Fig.\ref{fig:wros} demonstrates the sparse format and a microkernel execution example using WROS dataflow. The rotation of weights is performed offline, as illustrated in Fig.\ref{fig:wros-weight}, where weights for the output channel index $i$ are shifted within the weight block to the position $(i \bmod N)$. In WROS dataflow, only the weight data needs transformation, while the metadata (indices, indptr) remains unchanged. Fig.\ref{fig:wros-kernel} shows an example of performing operations using WROS dataflow with pre-rotated weights. In this configuration, when the starting output channel index of a tile is $i$, only the output vector corresponding to the $(i \bmod N)$-th row is stored in memory post-execution, while the rest of the output vector results are retained in the register file. For instance, in Tile 0, after $O_1$ is accumulated in register file $R_1$, Tile 1 can proceed with accumulations without the need for output register copying, as the necessary weight $D$ for calculating $O_1$ has already been rotated into $R_1$.

The UBP kernel with WROS dataflow is a transformation from the existing ABP kernel, involving two main changes. First, rather than storing all output vectors in memory after each tile's computation, as traditionally done, only one output vector per tile location is stored. Second, an epilogue code is added to store the remaining $N-1$ output vectors, which are still in the register file after the last tile operation, to memory. Initially, the implementation of the UBP kernel was developed based on ABP convolution kernels with a kernel size of 1 and stride of 1, utilizing kernels provided by XNNPACK. Other ABP convolution kernels were implemented using microkernels proposed by \textsc{nDirect}~\cite{ndirect}. Building upon these, we modified the ABP kernels to create the UBP kernel, which was integrated into XNNPACK to ensure end-to-end performance.

Moreover, to exploit multithreading parallelism, we partitioned the workload across the height dimension of the input data for both ABP and UBP kernel executions, assigning tasks to individual threads. This strategy ensures that all threads engage uniformly across the entire sparse weight matrix, avoiding issues like overlapped output tiles between threads and minimizing workload imbalance.

\section{Experiments}

\subsection{Experimental Setup}

We evaluated our methods on the ImageNet dataset using MobileNetV1 and ResNet50 models, training them on an RTX 3090 GPU. For comparison, we utilized the open-source SUBP~\cite{SUBP} implementation, a leading technique in the 1$\times$N pruning pattern. We followed the SUBP training hyper-parameters, referred to as Aligned Block-wise Pruning (ABP). However, for Unaligned Block-wise Pruning (UBP), we diverged by using the $\ell_1$ norm as the pruning criterion and implementing block regrowing at the element level based on importance scores, rather than at the block level as in SUBP. Despite these changes, UBP achieved higher accuracy than ABP.

The models were converted to TensorFlow Lite and tested on a Samsung Galaxy S20~\cite{galaxy_s20} with a Snapdragon 865, using ARM-v8.2-A ISA, an octa-core Qualcomm Kryo 585 CPU, and a Qualcomm Adreno 650 GPU. We measured execution times with a library of sparse kernels developed under our WROS dataflow, integrated into TensorFlow Lite using the TensorFlow Lite Model Benchmark Tool\footnote{https://github.com/tensorflow/tensorflow/tree/master/tensorflow/lite/tools/benchmark}.

\subsection{Performance of Block Selection Algorithm}

\begin{figure}[ht]
  \centering
  \begin{subfigure}{1.0\textwidth}
      \centering
      \includegraphics[width=\textwidth]{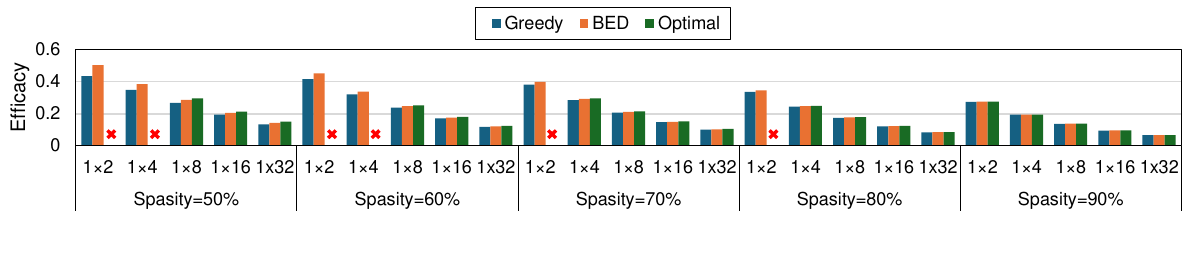}
      \vspace{-25pt}
      \caption{MobileNetV1}
      \label{fig:exp-efficacy-mbv1}
  \end{subfigure}
  
  \begin{subfigure}{1.0\textwidth}
      \centering
      \includegraphics[width=\textwidth]{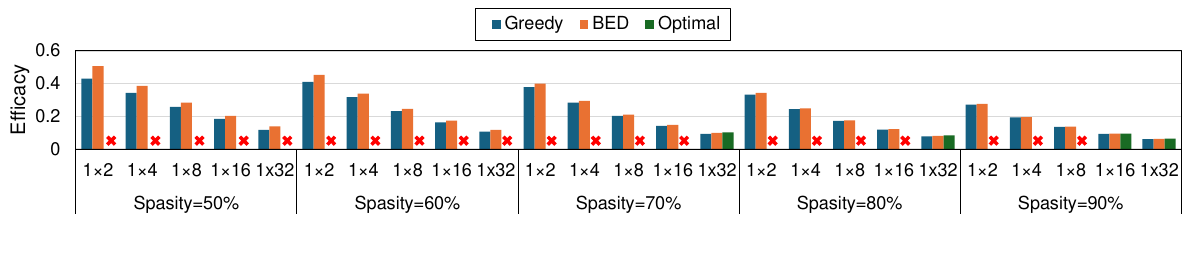}
      \vspace{-25pt}
      \caption{ResNet50}
      \label{fig:exp-efficacy-resnet50}
  \end{subfigure}
  \caption{Comparison of the efficacy of different block selection methods for UBP across various sparsity levels and block sizes}
  \label{fig:exp-efficacy}
  \vspace{-5pt}
\end{figure}

To evaluate the performance of block selection methods for Unaligned Block-wise Pruning (UBP), we defined an efficacy metric based on the $\ell_1$ norm. Specifically, the sum of the importance scores when performing element-wise pruning ($|W_{\text{EP}}|$) is set as 1.0 (maximum), and the sum when performing 1$\times$N aligned block-wise pruning ($|W_{\text{ABP(1}\times\text{N)}}|$) is set as 0.0 (minimum). This metric evaluates the total importance scores for each block selection method, with the efficacy for UBP (1$\times$N) defined as:
\begin{equation}
    efficacy = \frac{|W_{\text{UBP (1}\times\text{N)}}| - |W_{\text{ABP (1}\times\text{N)}}|}{|W_{\text{EP}}| - |W_{\text{ABP (1}\times\text{N)}}|}
\end{equation}

In Fig.\ref{fig:exp-efficacy}, we compare the efficacy of different block selection methods (greedy, VED, optimal) for UBP across all layers of MobileNetV1 and ResNet50, considering various sparsities and block sizes. The optimal method reports only partial experimental results because as block size and sparsity decrease, the required number of selected blocks increases, resulting in prohibitively long execution times. For MobileNetV1, this issue arises at sparsities of 80\% or less; for ResNet50, the challenge persists nearly across all sparsity levels when the block size is 8 or smaller.

In contrast, the Block Expansion and Division (BED) method demonstrated feasible execution times even for the most time-consuming layers of ResNet50, with a maximum duration of about 40 seconds for a block size of 2 and sparsity of 50\%. BED thus provides performance comparable to that of the optimal method while significantly reducing execution time, facilitating efficient training for UBP.

\subsection{Accuracy}

\begin{table}
\small
\caption{Accuracy comparison for ABP and UBP at varying sparsity levels in MobileNetV1 and ResNet50.}
\label{table:accuracy}
\centering
\begin{tabular}{c|cc|cc}
\hline
         & \multicolumn{2}{c|}{MobileNetV1 (baseline: 71.15)}         & \multicolumn{2}{c}{ResNet50 (baseline: 77.01)}              \\ \hline
Sparsity & \multicolumn{1}{c|}{ABP (1$\times$4)} & UBP (1$\times$4) & \multicolumn{1}{c|}{ABP (1$\times$4)} & UBP ($1\times$4)  \\ \hline
70\%     & \multicolumn{1}{c|}{71.126}   & 71.676   & \multicolumn{1}{c|}{76.466}   & 76.916    \\ \hline 
80\%     & \multicolumn{1}{c|}{69.652}   & 70.172   & \multicolumn{1}{c|}{76.018}   & 76.416    \\ \hline
90\%     & \multicolumn{1}{c|}{65.520}   & 66.836   & \multicolumn{1}{c|}{74.318}   & 74.684    \\ \hline
\end{tabular}
\vspace{-10pt}
\end{table}

Table~\ref{table:accuracy} presents the accuracy results for both Aligned Block-wise Pruning (ABP) and Unaligned Block-wise Pruning (UBP) applied to MobileNetV1 and ResNet50 across sparsity levels of 70\%, 80\%, and 90\%. Training for ABP utilized the SUBP method~\cite{SUBP}, which employs the BPAR pruning criterion. The results demonstrate that UBP consistently exhibits higher accuracy than ABP, a benefit attributable to the superior block selection method, suggesting that further enhancements could be achieved through a pruning criterion specifically tailored for UBP.

\subsection{UBP Kernel Performance}

\begin{figure}
  \centering
  \includegraphics[width=1.0\textwidth]{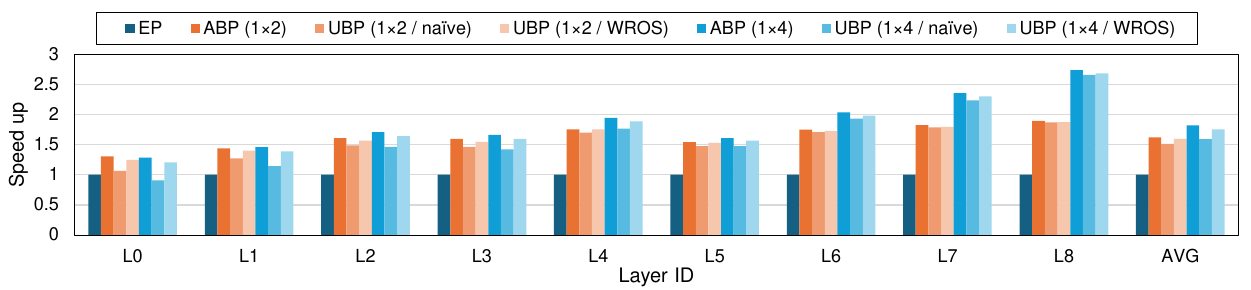}
  \caption{Comparison of UBP kernel performance using WROS dataflow across different layers of MobileNetV1.}
  \label{fig:exp-wros}
  \vspace{-10pt}
\end{figure}

In this subsection, we evaluate the performance of the UBP kernel with the Weight Rotating and Output Stationary (WROS) dataflow. Fig.\ref{fig:exp-wros} displays the relative performance of MobileNetV1 layers with a target sparsity of 80\%, comparing element-wise pruning (EP) and block-wise pruning (ABP and UBP) with block sizes of 2 and 4. EP and ABP performances were measured using XNNPACK's SpMM kernel, while UBP performance was evaluated using kernels implemented with both the naïve dataflow and our proposed WROS dataflow.

First, we observe that increasing block size in block-wise pruning enhances performance compared to EP due to improved data reuse in sparse operations. Secondly, the UBP kernel using WROS dataflow achieves performance nearly identical to the ABP kernel. The naïve dataflow implementation of UBP suffers from significant performance degradation due to required output register copying, which worsens with larger block sizes. In contrast, WROS dataflow eliminates unnecessary overhead by pre-rotating weights, allowing output register file values to remain unchanged. Thus, UBP with WROS dataflow achieves similar latency to ABP while offering higher accuracy, demonstrating its practical performance benefits.

\subsection{End-to-end Performance}

\begin{figure}
  \centering
  \begin{subfigure}{0.45\linewidth}
    \centering
    \includegraphics[width=\linewidth]{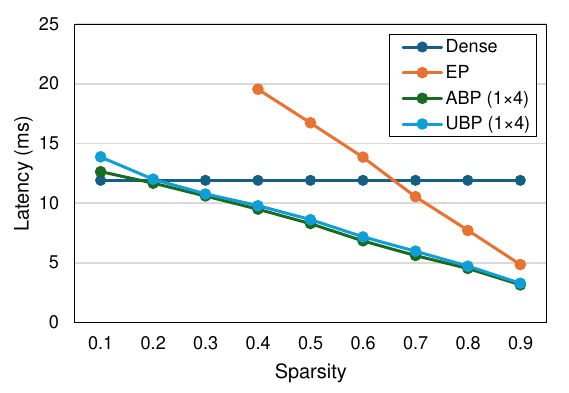}
    \caption{MobileNetV1}
    \vspace{-5pt}
    \label{fig:end-to-end-mbv1}
  \end{subfigure}
  \hfill
  \begin{subfigure}{0.45\linewidth}
    \centering
    \includegraphics[width=\linewidth]{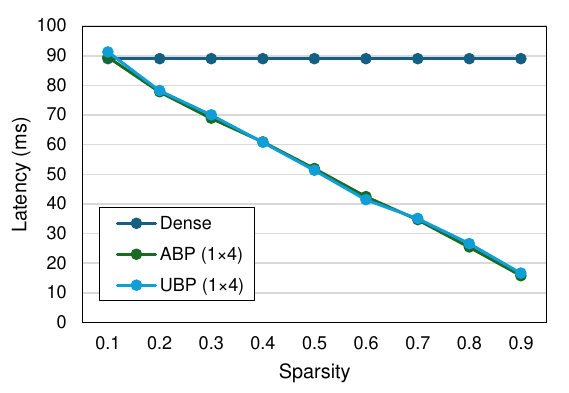}
    \caption{ResNet50}
    \vspace{-5pt}
    \label{fig:end-to-end-resnet50}
  \end{subfigure}
  \caption{End-to-end performance based on various sparsity levels and pruning patterns for MobileNetV1 and ResNet50.}
  \label{fig:end-to-end}
  \vspace{-10pt}
\end{figure}

Fig.\ref{fig:end-to-end} illustrates the latency of models with EP, ABP, and UBP at various sparsity levels on a Galaxy S20 CPU. The experiments utilized four threads for optimal performance. Notably, minimal performance differences between ABP and UBP were observed, attributed to the WROS dataflow minimizing overhead in the UBP kernel. For MobileNetV1, EP matches the performance of the dense model at approximately 65\% sparsity, whereas ABP and UBP achieve similar performance at 20\% sparsity due to enhanced kernel efficiency with larger block sizes. Similar trends were observed for ResNet50. At 70\% sparsity, where both MobileNetV1 and ResNet50 exhibit accuracy comparable to the dense model, UBP results in latency improvements of 2.0x and 2.5x, respectively, indicating that UBP is effectively optimized and suggests significant performance enhancements in practical applications.

\section{Conclusion}
This paper addresses the limited practical application of unaligned pruning due to training and inference challenges by introducing the Block Expansion and Division (BED) algorithm and the Weight Rotating and Output Stationary (WROS) dataflow. These innovations facilitate efficient training and reduce inference overhead, proving the practical viability of unaligned block-wise pruning. Our experiments demonstrate that Unaligned Block-wise Pruning (UBP) not only matches but exceeds the accuracy of leading block-wise pruning methods with comparable latency.

While UBP has shown promising results, it remains a relatively unexplored field. This paper focused primarily on adapting UBP for practical use rather than proposing new pruning criteria or training methods. We have demonstrated that UBP can outperform aligned block-wise pruning under certain conditions. Looking forward, we anticipate that further research will introduce more refined techniques specifically designed for UBP, potentially unlocking even greater improvements in performance.

\bibliographystyle{IEEEtran}
\bibliography{ref}

\end{document}